\documentclass[sigconf]{acmart}

\usepackage{algorithm}
\usepackage{algorithmic}
\usepackage{amsthm}
\usepackage{amsmath}
\usepackage{bm}
\usepackage{dsfont}
\usepackage{multirow}
\usepackage{color}
\usepackage{subcaption}
\usepackage{booktabs}

\newtheorem{theorem}{Theorem}

\AtBeginDocument{
  }

\copyrightyear{2025}
\acmYear{2025}
\setcopyright{rightsretained}
\acmConference[CIKM '25] {Proceedings of the 34th ACM International Conference on Information and Knowledge Management}{ November 10--14, 2025}{Seoul, Republic of Korea.}
\acmBooktitle{Proceedings of the 34th ACM International Conference on Information and Knowledge Management (CIKM '25), November 10--14, 2025, Seoul, Republic of Korea}
\acmISBN{979-8-4007-2040-6/2025/11}
\acmDOI{10.1145/XXXXXX.XXXXXX}

\begin{document}

\title{Towards Unveiling Predictive Uncertainty Vulnerabilities in the Context of the Right to Be Forgotten}

\author{Wei Qian}
\authornote{The first two authors contribute equally to this work.}
\affiliation{
  \institution{Iowa State University}
  \city{Ames}
  \state{Iowa}
  \country{USA}
}
\email{wqi@iastate.edu}

\author{Chenxu Zhao}
\authornotemark[1]
\affiliation{
  \institution{Iowa State University}
  \city{Ames}
  \state{Iowa}
  \country{USA}
}
\email{cxzhao@iastate.edu}

\author{Yangyi Li}
\affiliation{
  \institution{Iowa State University}
  \city{Ames}
  \state{Iowa}
  \country{USA}}
\email{liyangyi@iastate.edu}

\author{Wenqian Ye}
\affiliation{
  \institution{University of Virginia}
  \city{Charlottesville}
  \state{Virginia}
  \country{USA}}
\email{wenqian@virginia.edu}

\author{Mengdi Huai}
\affiliation{
  \institution{Iowa State University}
  \city{Ames}
  \state{Iowa}
  \country{USA}}
\email{mdhuai@iastate.edu}

\renewcommand{\shortauthors}{Wei Qian, Chenxu Zhao, Yangyi Li, Wenqian Ye, and Mengdi Huai}

\begin{abstract}
Currently, various uncertainty quantification methods have been proposed to provide certainty and probability estimates for deep learning models' label predictions. Meanwhile, with the growing demand for the right to be forgotten, machine unlearning has been extensively studied as a means to remove the impact of requested sensitive data from a pre-trained model without retraining the model from scratch. However, the vulnerabilities of such generated predictive uncertainties with regard to dedicated malicious unlearning attacks remain unexplored. To bridge this gap, for the first time, we propose a new class of malicious unlearning attacks against predictive uncertainties, where the adversary aims to cause the desired manipulations of specific predictive uncertainty results. We also design novel optimization frameworks for our attacks and conduct extensive experiments, including black-box scenarios. Notably, our extensive experiments show that our attacks are more effective in manipulating predictive uncertainties than traditional attacks that focus on label misclassifications, and existing defenses against conventional attacks are ineffective against our attacks.
\end{abstract}



\maketitle

\section{Introduction}
\label{sec:intro}
Deep neural networks (DNNs) have emerged as powerful tools with a profound impact on various real-world tasks \cite{wang2024limaml, yang2023pyhealth,wei2024process, jiang2021dl, qiu2018deepinf,yang2018deep,boone2022marine}. Nevertheless, the direct adoption of outcomes from these opaque models often raises apprehension. Uncertainty quantification (UQ) has emerged as an essential strategy to enhance the robustness and reliability of deep learning models by providing a quantitative measure of confidence in model predictions.

Currently, numerous uncertainty quantification methods \cite{kong2023uncertainty,chen2024modeling,gal2016dropout, romano2020classification,qian2024towards,angelopoulos2020uncertainty} have been proposed to offer insights into model confidence and aid researchers to enhance model performance \cite{qiao2021uncertainty,ji2023uncertainty,yin2024sepsislab,lidder2025neuron}. They can be generally divided into two categories. The first category involves methods that do not modify the underlying model but act as wrappers in a post-hoc setting. For instance, softmax methods \cite{gawlikowski2023survey} and deep ensembles \cite{lakshminarayanan2017simple} fall under this category, where they do not alter the model's structure. These methods are applied after the model has been trained and are used to enhance the interpretability and reliability of predictions. The second category requires modifications to the model architecture itself. Bayesian neural networks (BNNs) \cite{tran2019bayesian} exemplify this category, where probability distributions are imposed over model parameters to provide a comprehensive representation of model uncertainty.

On the other hand, with the increase of the right to be forgotten, machine unlearning \cite{bourtoul2021,qian2022patient,warnecke2023machine,chen2025survey} has been extensively researched in recent years. Many privacy protection regulations, such as the European Union’s General Data Protection Regulation (GDPR) \cite{regulation2016regulation} and the California Consumer Privacy Act \cite{pardau2018california}, mandate that data providers can request deletion of their private data. Machine unlearning enables the removal of such data from pre-trained DNNs without the extremely high cost of retraining from scratch~\cite{bourtoul2021}.

However, in practice, the increasingly critical role of machine unlearning makes models' predictive uncertainty susceptible to the risk of being maliciously attacked. Traditional unlearning attacks \cite{liu2024backdoor,qian2023towards,zhao2024static,hu2024duty, zhao2024rethinking,huang2024uba} primarily focus on causing label misclassifications, while our focus is on investigating vulnerabilities related to prediction uncertainties. Malicious unlearning attacks targeting predictive uncertainty can be more subtle and harder to detect, as these attacks manipulate predictive uncertainty without directly altering the original predicted labels, potentially bypassing existing defenses that rely on detecting label changes. Moreover, directly applying traditional unlearning attacks in this predictive uncertainty setting suffers from limitations in stealthiness and efficiency.

In this work, we perform the first study on malicious unlearning attacks against predictive uncertainty, where the adversary aims to undermine the use of uncertainty quantification techniques by manipulating the generated predictive uncertainties while \emph{preserving label predictions}. While there are a few existing works \cite{zeng2022attacking,obadinma2024calibration,zeng2023manipulating,li2024data} studying security risks on uncertainty quantification, they do not consider the risks of malicious unlearning attacks during the unlearning process. The main challenge here is how to ensure the stealthiness of the performed malicious unlearning attacks. Studies of DNNs’ robustness have enabled advances in defending against adversarial attacks and poisoning attacks. However, existing defenses \cite{yu2022robust,wang2023better,xu2021adversarial,wang2023mix-up,shi2024adversarial} are often effective only against a specific attacking type of traditional adversarial and poisoning attacks, drastically degrade the generalization performance, or are computationally prohibitive for standard machine unlearning pipelines.

Motivated by the above, we thus believe that studying unlearning attacks targeting predictive uncertainty is crucial for safety applications of existing uncertainty quantification methods. This work makes the following contributions: For the first time, we study the vulnerability and robustness of uncertainty to unlearning attacks. To this end, we design a novel unlearning attack framework to craft effective unlearning requests against predictive uncertainties. Notably, our method is grounded in empirical findings, which are further supported by rigorous theoretical analysis. Specifically, in our method, to ensure attack stealthiness, we first design a regularized predictive uncertainty attack loss to make target samples appear more natural within the data distribution. Additionally, we present novel efficient optimization methods by rigorously refining our proposed attacks of generating effective malicious unlearning requests through the closed-form updates, thus enhancing the efficiency of the proposed attacks. We perform extensive experiments to validate the desired properties of our attacks across various scenarios (e.g., black-box settings). Our experiments also examine existing defenses against traditional attacks, revealing that they are insufficient in countering our attacks. This underscores the need for new strategies to address these advanced unlearning threats and highlights the potential negative impacts of unlearning attacks.

\section{Methodology}
\label{sec:method}
\textbf{Threat model.} We consider a scenario where a model owner has a well-trained model based on a training dataset. An adversary pretends to be the normal data provider, and aims to initiate malicious unlearning requests to delete certain data during the unlearning process, such that the predictive uncertainties of the unlearned model will be compromised. We consider both \emph{white-box} and \emph{black-box} settings. In the white-box setting, the adversary has complete knowledge. This scenario is crucial for examining the strongest attacking behavior~\cite{neekhara2021adversarial, chen2020frank, qian2023towards, liu2024backdoor}. In the black-box setting, the adversary has no prior knowledge of the threat model. The adversary aims to perform two forms of unlearning attacks to manipulate predictive uncertainties: \emph{overconfidence} and \emph{underconfidence}. Specifically, the overconfidence attack increases prediction confidence, causing the model to be overconfident in its predictions, while the underconfidence attack decreases prediction confidence, resulting in the model being underconfident in its predictions.

\begin{table*}[t!]
\small
\centering
\caption{Attack performance in terms of ECE on various uncertainty quantification methods.}
\label{tab:attack_uncertainty_quantification}
\vskip -10pt
\begin{tabular}{cccccccccc}
\hline
\multirow{2}{*}{Method} & \multicolumn{3}{c}{CIFAR-10} & \multicolumn{3}{c}{CIFAR-100} & \multicolumn{3}{c}{ImageNet-100} \\
\cline{2-10}
& Initial & Rand &  \textbf{Ours} & Initial & Rand &  \textbf{Ours} & Initial & Rand & \textbf{Ours} \\
\hline
\multirow{1}{*}{Softmax} & $.100_{\pm .018}$ & $.100_{\pm .018}$ & $\bm{.338_{\pm .066}}$ & $.204_{\pm .026}$ & $.209_{\pm .024}$ & $\bm{.431_{\pm .038}}$ & $.139_{\pm .011}$ & $.191_{\pm .013}$ & $\bm{.353_{\pm .017}}$ \\
\multirow{1}{*}{Deep ensemble} & $.115_{\pm .018}$ & $.115_{\pm .018}$ & $\bm{.287_{\pm .063}}$  & $.204_{\pm .017}$ & $.207_{\pm .018}$ & $\bm{.391_{\pm .035}}$ & $.153_{\pm .021}$ & $.151_{\pm .025}$ & $\bm{.255_{\pm .034}}$ \\
\multirow{1}{*}{MC dropout} & $.117_{\pm .015}$ & $.107_{\pm .017}$ & $\bm{.313_{\pm .065}}$ & $.199_{\pm .024}$ & $.198_{\pm .027}$ & $\bm{.363_{\pm .031}}$ & $.229_{\pm .020}$ & $.240_{\pm .023}$ & $\bm{.291_{\pm .021}}$ \\
\multirow{1}{*}{TS} & $.138_{\pm .013}$ & $.138_{\pm .013}$ & $\bm{.242_{\pm .022}}$ & $.215_{\pm .014}$ & $.206_{\pm .019}$ & $\bm{.293_{\pm .029}}$ & $.157_{\pm .016}$ & $.199_{\pm .016}$ & $\bm{.347_{\pm .030}}$ \\
\multirow{1}{*}{ETS} & $.139_{\pm .012}$ & $.138_{\pm .012}$ & $\bm{.278_{\pm .040}}$ & $.199_{\pm .017}$ & $.212_{\pm .016}$ & $\bm{.301_{\pm .025}}$ & $.178_{\pm .018}$ & $.212_{\pm .016}$ & $\bm{.345_{\pm .017}}$ \\
\multirow{1}{*}{IR} & $.096_{\pm .016}$ & $.097_{\pm .017}$ & $\bm{.206_{\pm .027}}$ & $.209_{\pm .038}$ & $.206_{\pm .029}$ & $\bm{.282_{\pm .028}}$  & $.157_{\pm .015}$ & $.143_{\pm .019}$ & $\bm{.190_{\pm .020}}$ \\
\hline
\end{tabular}
\vskip -2pt
\end{table*}

\textbf{Attack formulation.} Here, we present the idea of our proposed attack in the underconfidence setting. Let \(\mathcal{E}: \mathcal{X} \times \Theta \rightarrow\) \(\mathcal{P}(\mathcal{Y})\) denote an uncertainty estimation method, where \(\mathcal{X}\) is the data space, \(\Theta\) represents the space of model parameters, and \(\mathcal{P}(\mathcal{Y})\) denotes the space of probability distributions over the set of labels \(\mathcal{Y}\). For each pair \((x, \theta) \in \mathcal{X} \times \Theta\), where $x$ is an input and \(\theta\) is the model parameter, \(\mathcal{E}(x; \theta)\) outputs a probability distribution on \(\mathcal{Y}\) that quantifies the uncertainty of label predictions for \(x\) under the parameter \(\theta\). Then, we can define \(\hat{y}\) as the label corresponding to the maximum probability estimated by the uncertainty method \(\mathcal{E}\) for the input \(x\). Formally, \(\hat{y}\) can be defined as \(\hat{y}=\arg \max _{c \in \mathcal{Y}} \mathcal{E}^{c}(x; \theta)\), where \(\mathcal{E}^{c}(x; \theta)\) denotes the probability that \(x\) belongs to label \(c\). For example, for softmax uncertainty estimation method, the prediction \(\hat{y}\) simplifies to the ordinary label prediction, i.e., \( \hat{y} = \arg\max_{c \in \mathcal{Y}} F_{L}^{c}(x; \theta) \), where \(L\) is the softmax layer in model \(F(\theta)\). Let \( \theta^* \) denote a pre-trained model, based on the training dataset \( \mathcal{D} \) and a learning algorithm \( \mathcal{A} \), and \( \mathcal{D}_{t}=\{(x_t, y_t)\}_{t=1}^{T} \subset \mathcal{D} \) be a subset accessible to the adversary. We introduce a discrete indication weight \( w_{t} \in \{0,1\} \) to specify whether the sample \( x_{t} \) should be removed (\( w_{t}=1 \)) or retained (\( w_{t}=0 \)), forming the forget set \( \mathcal{D}_{u}=\mathcal{D}_{t}\circ \Phi=\{(x_{t}, y_{t})| (x_{t}, y_{t}) \in \mathcal{D}_{t} \textnormal{ and } w_{t}=1\}\), with \( \Phi=\{w_{t} \}_{t=1}^{T} \). Let \( \mathcal{D}_v = \{(x_v, y_v)\}_{v=1}^{V} \) denote a set of victim target samples, the adversary's goal is to generate a forget set \( \mathcal{D}_u \) (i.e., the set of indication weights \( \Phi=\{w_t\}_{t=1}^{T} \)) and derive an unlearned model \( \theta^u \) using an unlearning algorithm \( \mathcal{U} \), such that the unlearned model being underconfident in prediction confidence of the target samples. To achieve this goal, we formulate the following underconfidence attack loss
\begin{align}
\label{eq:underconfidence}
     &  \ell_1(\theta^u; \mathcal{D}_v) = \sum_{v=1}^{V} \max (\mathcal{E}^{\hat{y}_v}(x_v; \theta^u) - \max_{c \neq \hat{y}_v} \mathcal{E}^{c}(x_v; \theta^u), 0), 
\end{align}
\noindent where \( \theta^u = \mathcal{U}(\theta^*, \mathcal{D}, \mathcal{D}_u=\mathcal{D}_t \circ \Phi) \) is the unlearned model, \( \mathcal{E}^{\hat{y}_v}(x_v; \theta^u) \) represents the maximized prediction probability for the target sample \( x_v \), and \( \max(\cdot, 0) \) is the hinge loss.

\begin{figure}[htbp]
\vskip -6pt
\centering
\includegraphics[width=0.95\linewidth]{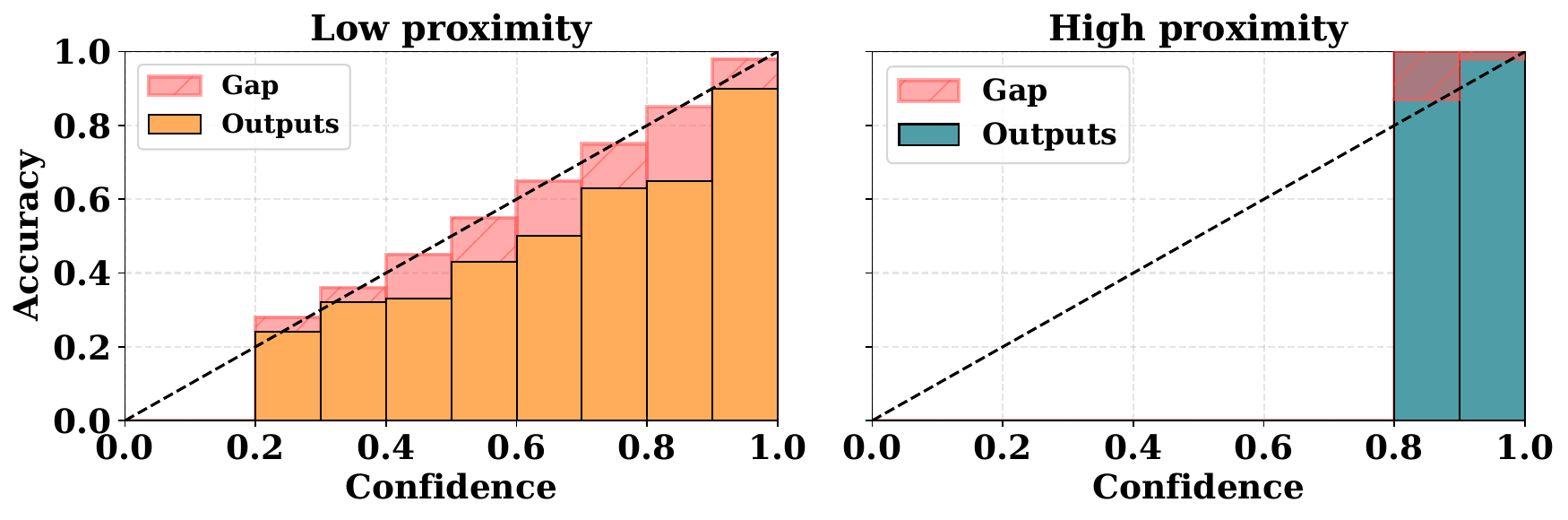}
\vskip -12pt
\caption{Relationship between the proximity and confidence on CIFAR-10. Samples with higher (lower) proximity tend to be more underconfident (overconfident).}
\label{fig:proximity}
\vskip -6pt
\end{figure}

However, when optimizing the formulated loss in Eq.~(\ref{eq:underconfidence}) to craft unlearning samples, the uncertainty distribution of the victim target samples might differ from those of other underconfidence data in the distribution. This discrepancy can make the attack suspicious and increase the risk of being detected. This is also observed in our experiments. In Figure~\ref{fig:proximity}, we present two examples to show the relationship between proximity and confidence. From this figure, we can see that the samples with varying proximities tend to exhibit different levels of prediction confidence. Typically, the low-proximity samples are more prone to being overconfident, whereas high-proximity samples are more likely to be underconfident. Theorem \ref{thm:proximity} further supports these empirical findings, showing that the effect of overconfidence diminishes as the number of training data increases. The proof of this theorem is deferred to the full version.
\begin{theorem}
\label{thm:proximity}
    Consider a data generation process where \( X \) is standard Gaussian and \( Y | X \) follows a binary model with activation function \( \sigma (x)= 1 - \frac{1}{1+e^{-x}} \), i.e. \( X \sim \mathsf{N}\left(0, \mathrm{I}_d\right), \mathbb{P}(Y=1 \mid X=x)=\sigma\left(\theta_{\star}^{\top} x\right) \), where $\mathbf{\theta}_{\star}$ is the ground truth coefficient vector.  If we obtain the minimizer \( \hat{\theta} \) for n training data through empirical risk minimization, we can deduce the following result: in the limit as \( n, d \rightarrow \infty \) and \( d / n \rightarrow \kappa \) for small enough \( \kappa>0 \), for any \( p=\sigma(\hat{\theta}^{\top} x)\in(0.5,1) \), almost surely we have \(\Delta_p^{\mathrm{cal}} \rightarrow \mathcal{G}_{p,\kappa} = \mathcal{G}_p \cdot \kappa+o(\kappa) \quad \text{for some } \mathcal{G}_p > 0\), where \(\Delta_p^{\mathrm{cal}}\) is the calibration error for the binary model.
\end{theorem}

To address the above challenge, we design a novel proximity-based regularizer, which can achieve a natural uncertainty distribution for our proposed unlearning attacks. To capture the proximity of a sample  \( x \) \cite{xiong2023proximity}, we calculate the average distance between \( x \) and its \( K \)-nearest neighbors \( \mathcal{N}_{K}(x) \) in the data distribution
\begin{align}
\label{eq:proximity}
   & G_{PR}(x) = \exp (- \frac{1}{K} \sum_{x_i \in \mathcal{N}_K(x)} G_{EU} (F_{H}(x; \theta), F_{H}(x_i; \theta) ) ), 
\end{align}
\noindent where \( G_{EU} \) is the Euclidean distance function, and \(H\) represents a hidden layer. In the above, \( G_{PR}(x) \) captures the local density of a sample and its relationship to its neighborhood. To align the uncertainty distribution of target samples with those of high-proximity data, for every target sample \( x_v \), we first formulate a set \(\mathcal{D}_p^v\) for high-proximity data based on Eq.~(\ref{eq:proximity}) within the same predicted label. Then, we propose to optimize the uncertainty distribution by measuring the KL divergence of the uncertainty vectors between the target sample and the high-proximity data. Then, we formulate
\begin{align}
   \label{eq:underconfidence2}
   & \ell_2(\theta^u; \mathcal{D}_v) =  \sum_{v=1}^{V} G_{KL}(\mathcal{E}(x_v; \theta^{u}) || \frac{1}{|\mathcal{D}_p^v|}\sum_{(x_i, y_i) \in \mathcal{D}_p^v} \mathcal{E}(x_i; \theta^{u})),  
\end{align}
\noindent where \( G_{KL}(\cdot||\cdot) \) denotes the KL divergence measure, and \( \mathcal{D}_p^v = \{(x_i,y_i) \in \mathcal{D}' \mid y_i = \hat{y}_v \wedge G_{PR}(x_i) \geq \Xi\} \) represents the set of high-proximity samples for \( x_v \), with \( \Xi \) being an approximate threshold and \( \mathcal{D}' \) is a holdout dataset (e.g., a validation set). This regularizer aligns the uncertainty vectors of target samples with those of the high-proximity data, enabling the adversary to achieve natural uncertainty distribution patterns and enhance attack stealthiness.

Based on the above, to effectively manipulate the predictive uncertainty of the victim target samples \(\mathcal{D}_v\), we formulate 
\begin{align}
\label{eq:undercondifence4_opt}
    & \min_{\Phi} \ell_3(\theta^u; \mathcal{D}_v) = \ell_1(\theta^u; \mathcal{D}_v)+ \lambda \ell_2(\theta^u; \mathcal{D}_v) \\
    & \quad s.t. \quad \theta^{u} =  \mathcal{U}(\theta^{*}, \mathcal{D}, \mathcal{D}_u = \mathcal{D}_t \circ \Phi),\notag 
\end{align}
\noindent where \( \lambda \) is a hyperparameter. In the above, \( \ell_1(\theta^u; \mathcal{D}_v) \) is the underconfidence attack loss in Eq. (\ref{eq:underconfidence}), and \(  \ell_2(\theta^u; \mathcal{D}_v) \) is the regularizer in Eq. (\ref{eq:underconfidence2}) that achieves natural uncertainty distributions.

\textbf{Optimization.} Note that the attack framework described in Eq.~(\ref{eq:undercondifence4_opt}) is a bi-level optimization problem, where the outer optimization defines the unlearning attack objective and the inner problem specifies the model's unlearning objective. However, directly solving this bi-level optimization can be computationally expensive. To address this, we approximate the loss \(\ell_3(\theta^u; \mathcal{D}_v) \) as follows
\begin{align}
    & \quad \min_{\Phi} \ell_3(\theta^u; \mathcal{D}_v)  = \min_{\Phi} \ell_3(\theta^u; \mathcal{D}_v) - \ell_3(\theta^*; \mathcal{D}_v) \\
    & \approx \min_{\Phi} \nabla_\theta \ell_3(\theta^*; \mathcal{D}_v)^T (\theta^u - \theta^*) = \min_{\Phi} - \tau \nabla_\theta \ell_3(\theta^* ;\mathcal{D}_v)^T \Psi( \theta^*,\mathcal{U}), \notag
\end{align}
\noindent where \(\Psi( \theta^*,\mathcal{U})\) is the model update function when unlearning method \(\mathcal{U}\) is applied to the original model parameters \(\theta^*\), and \( \tau \) is the update’s step size. Hence, to induce a model update \(\Psi(\theta^*,\mathcal{U})\) that maximizes the increase in loss for \(\mathcal{D}_v\), we seek to minimize the above equation for small values of parameter \(\tau\). Thus, our objective transforms into solving the following gradient alignment problem
\begin{align}
\max_{\Phi} \frac{\nabla_\theta \ell_3(\theta^* ; \mathcal{D}_v)^T \Psi(\theta^*,\mathcal{U})}{\| \nabla_\theta \ell_3(\theta^*; \mathcal{D}_v)\| \| \Psi(\theta^*,\mathcal{U})\|}.
\label{eq:gradient_match}
\end{align}
\noindent The above optimization aligns the directions of \( \nabla_\theta \ell_3(\theta^*;\mathcal{D}_v) \) and \( \Psi(\theta^*,\mathcal{U}) \) to achieve the maximal attack effect. In practice, we relax the indication weights to a continuous range \( [0, 1] \) for optimization, and incorporate random restarts to minimize the loss multiple times, beginning with random initial indication weights and selecting the optimal one to improve the reliability of the attack process.

\textbf{Discussion.} For the threat of malicious unlearning to predictive uncertainty, we also consider the black-box setting, where the adversary can train several surrogate models to generate malicious unlearning requests, along with alternative unlearning
methods. These requests can be transferred to the target model by exploiting the transferability ~\cite{wang2019transferable, wang2021admix, schwarzschild2021just}. This allows the adversary to execute malicious unlearning attacks in the black-box scenario effectively.

\section{Experiments}
\label{sec:exp}

\subsection{Experimental Setup}

\textbf{The adopted uncertainty quantification methods.} In experiments, we adopt diverse uncertainty quantification methods. In uncertainty estimation, we adopt MC dropout ~\cite{gal2016dropout} (the proxy to BNNs), as well as the softmax score ~\cite{gawlikowski2023survey} and deep ensemble~\cite{lakshminarayanan2017simple}. We also consider several uncertainty calibration techniques, including isotonic regression (IR)~\cite{zadrozny2002transforming}, temperature scaling (TS)~\cite{guo2017calibration}, and ensemble TS (ETS)~\cite{zhang2020mix}. We further adopt the following conformal prediction methods: HPS \cite{lei2013distribution}, APS \cite{romano2020classification}, and RAPS \cite{angelopoulos2020uncertainty}.

\textbf{The adopted machine unlearning methods.} In experiments, we adopt several popular unlearning methods, including the first-order based unlearning~\cite{warnecke2023machine}, unrolling SGD~\cite{thudi2022unrolling}, Fisher forgetting~\cite{golatkar2020eternal}, second-order based unlearning~\cite{warnecke2023machine}, selective synaptic dampening (SSD)~\cite{foster2024fast}, and SISA~\cite{bourtoul2021}. For each experiment, a random set of victim target samples is selected for attack, with each sample’s available training data drawn from its \( K \)-nearest neighbors.

\begin{figure}[htbp]
\vskip -6pt
\centering
\begin{subfigure}{0.465\linewidth}
\includegraphics[width=1.0\linewidth]{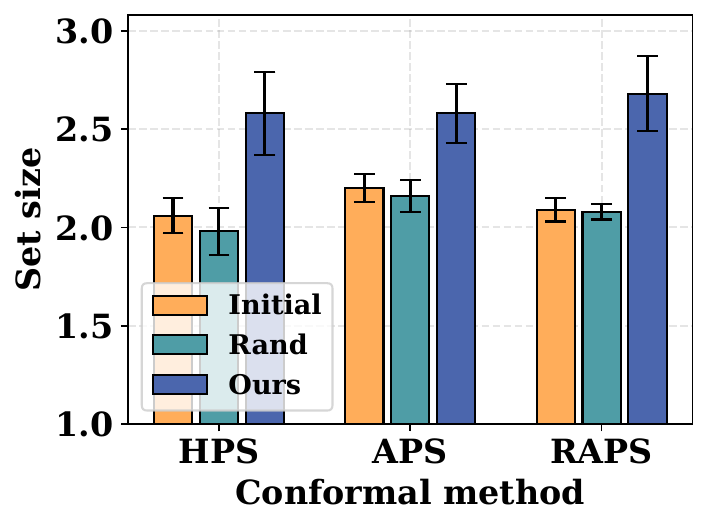}
\vskip -8pt
\caption{Attack performance}
\label{fig:performance_cp}
\end{subfigure}
\begin{subfigure}{0.445\linewidth}
\includegraphics[width=1.0\linewidth]{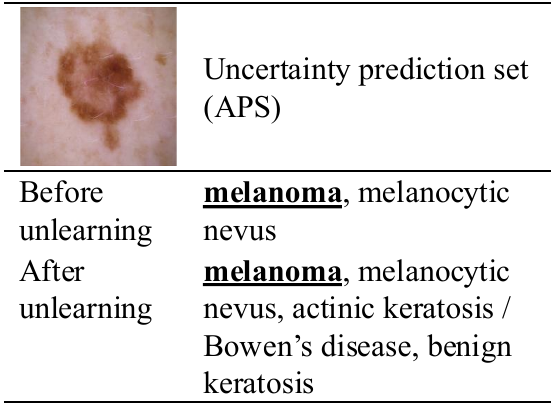}
\vskip -4pt
\caption{Visualization}
\label{fig:visual_cp}
\end{subfigure}
\vskip -12pt
\caption{Attack performance on conformal prediction.}
\label{fig:attack_cp}
\vskip -8pt
\end{figure}

\textbf{Datasets, models, and baselines.} In experiments, we adopt the following datasets: ISIC 2018~\cite{codella2019skin}, ImageNet-100~\cite{deng2009imagenet}, CIFAR-100~\cite{cifar-10}, and CIFAR-10~\cite{cifar-10}. We also employ various deep learning models, including ResNet-18~\cite{he2016deep}, VGG-19~\cite{simonyan2014very}, and MobileNetV2~\cite{sandler2018mobilenetv2}. Since there are no existing unlearning attacks specifically attacking uncertainty quantification, we first adopt a \emph{Rand} baseline that removes random samples from the training data. Additionally, to contrast with traditional attacks that aim to induce label misclassifications, we adopt the proposed method in \cite{zhao2024static}, which seeks to cause label misclassifications via manipulated data removal requests.

\begin{figure}[htbp]
\vskip -6pt
\centering
\begin{subfigure}{0.455\linewidth}
\includegraphics[width=1\linewidth]{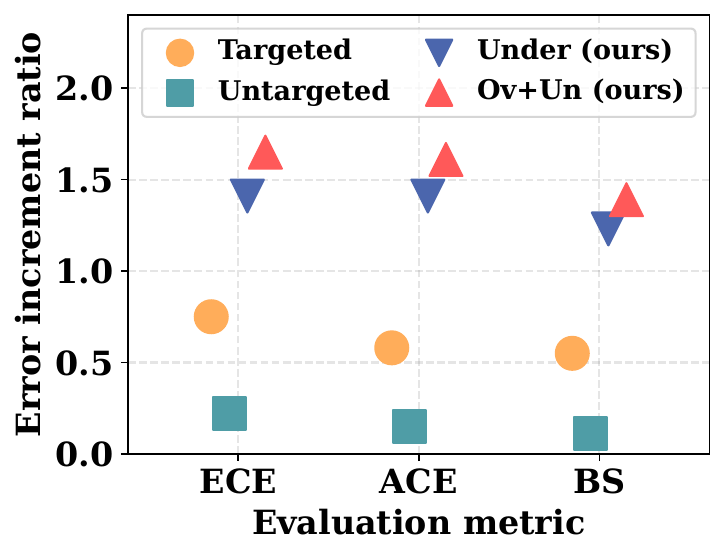}
\vskip -8pt
\caption{Attack comparison}
\label{fig:attack_compare}
\end{subfigure}
\begin{subfigure}{0.455\linewidth}
\includegraphics[width=1\linewidth]{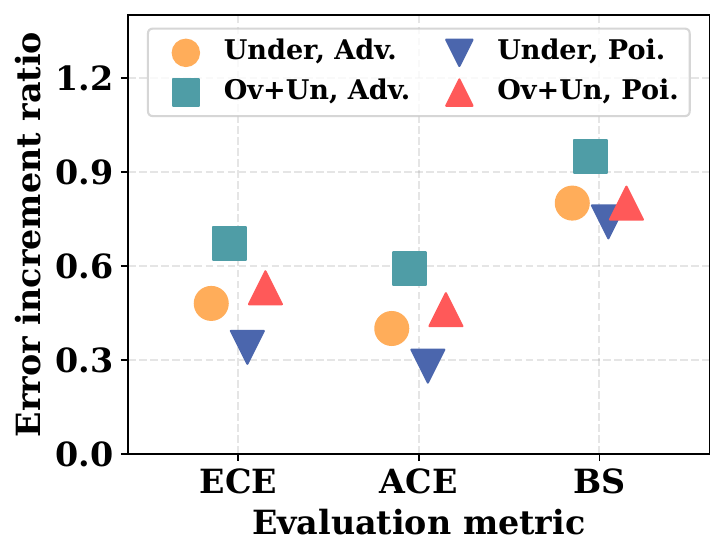}
\vskip -8pt
\caption{Defense comparison}
\label{fig:defense_compare}
\end{subfigure}
\vskip -12pt
\caption{Performance comparison of traditional attacks and existing defenses on CIFAR-10.}
\label{fig:attack_defense}
\vskip -12pt
\end{figure}

\subsection{Experimental Results}

\textbf{Attack performance on uncertainty quantification.} We first present the performance of our proposed unlearning attacks on various uncertainty quantification methods. Besides overconfidence (\emph{Over}) and underconfidence (\emph{Under}) unlearning attacks, we also experiment with a combined attack (\emph{Ov+Un}) that considers both overconfidence and underconfidence to produce maximum uncertainty errors. We evaluate the uncertainty errors using metrics such as Expected Calibration Error (ECE)~\cite{naeini2015obtaining}, Adaptive Calibration Error (ACE)~\cite{nixon2019measuring}, and Brier Score (BS)~\cite{brier1950verification}. Table \ref{tab:attack_uncertainty_quantification} highlights the performance of our combined attack. We use ResNet-18 for training and adopt the first-order based method for unlearning. The experimental results show that our proposed attacks significantly outperform the random baseline across all uncertainty quantification methods. Furthermore, Figure~\ref{fig:attack_cp} presents quantitative and qualitative results of the underconfidence attack on conformal prediction using the ISIC 2018 medical image dataset. The underlined and bold phrases represent true labels. Thus, our proposed unlearning attacks are highly effective in disrupting predictive uncertainty.

\textbf{Comparison with traditional attacks and defenses.} Then, Figure~\ref{fig:attack_compare} compares our attacks with traditional unlearning attacks in targeted and untargeted settings~\cite{zhao2024static} on uncertainty. We measure the error increment ratios across different evaluation metrics on the label-preserved data for each attack scenario, using first-order based unlearning. As shown, our attacks significantly increase uncertainty errors by specifically targeting prediction confidence, in contrast to traditional unlearning attacks. Figure~\ref{fig:defense_compare} further shows that our attacks remain effective even under existing defenses like adversarial training~\cite{madry2017towards} (\emph{Adv.}) and adversarial poisoning~\cite{geiping2021doesn} (\emph{Poi.}).

\begin{figure}[htbp]
\vskip -10pt
\centering
\begin{subfigure}{0.475\linewidth}
\includegraphics[width=0.92\linewidth]{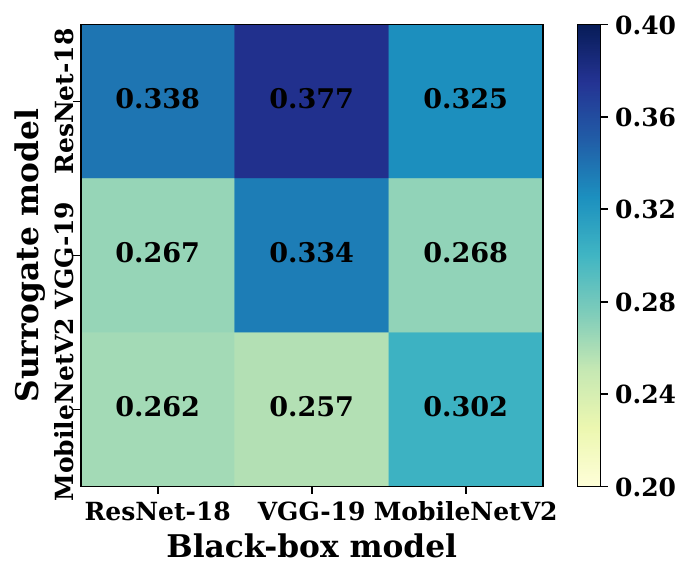}
\vskip -8pt
\caption{Across model architecture}
\label{fig:blackbox_model}
\end{subfigure}
\begin{subfigure}{0.475\linewidth}
\includegraphics[width=0.985\linewidth]{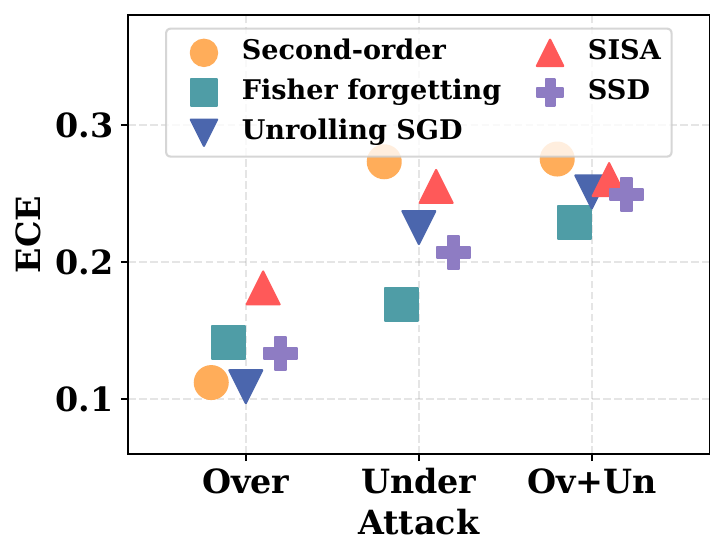}
\vskip -8pt
\caption{Arcoss unlearning method}
\label{fig:blackbox_unlearn}
\end{subfigure}
\vskip -12pt
\caption{Transferability of unlearning attacks.}
\label{fig:more_exp}
\vskip -10pt
\end{figure}

\textbf{Black-box setting.} At last, we investigate the performance of our unlearning attacks in the black-box setting, using ECE on CIFAR-10. Figure~\ref{fig:blackbox_model} shows that our attacks successfully transfer the unlearning data to attack the black-box model, even though it is trained with a different architecture than the surrogate model. Figure~\ref{fig:blackbox_unlearn} demonstrates successful transferability across unlearning methods. The unlearning data crafted using first-order based unlearning can be effectively transferred to SISA, SSD, unrolling SGD, Fisher forgetting, and second-order based unlearning methods to disrupt uncertainty. These results underscore the effectiveness of our unlearning attacks in the black-box setting.

\section{Conclusion}
\label{sec:conclusion}
To the best of our knowledge, we are the first to investigate the vulnerability and robustness of predictive uncertainty to malicious unlearning attacks during the unlearning process within the context of machine unlearning. To this end, we devise a novel attack framework for crafting effective unlearning requests against predictive uncertainty. Specifically, we first design a regularized predictive uncertainty attack loss, which not only preserves predicted labels but also makes the victim samples appear more natural within the data distribution. Further, we present novel and efficient optimization methods by rigorously refining our attacks of generating effective malicious unlearning requests through closed-form model updates. We perform theoretical analysis for our attacks. Extensive experiments verify the effectiveness of our proposed attacks.

\begin{acks}
This work is supported in part by the US National Science Foundation under grants CNS-2350332 and IIS-2442750. Any opinions, findings, and conclusions or recommendations expressed in this material are those of the author(s) and do not necessarily reflect the views of the National Science Foundation.
\end{acks}

\section*{GenAI Usage Disclosure}
During manuscript preparation, we utilized generative AI tools such as ChatGPT-4o solely for editorial purposes, for example,  checking grammar,  correcting spelling, and improving readability. The AI model was not used to generate scientific ideas, perform analyses, draw conclusions, or write code.

\bibliographystyle{ACM-Reference-Format}
\balance
\bibliography{ref}

\end{document}